\documentclass{llncs}

\usepackage[utf8]{inputenc}
\usepackage{mathpazo}
\usepackage{amsmath}
\usepackage{hyperref}
\usepackage[dvipsnames]{xcolor}
\usepackage{graphicx}
\usepackage{cleveref}

\usepackage{comment}

\newcommand{\diff}{\mathrm{d}}

\title{A function space perspective on stochastic shape evolution}
\author{Elizabeth Baker\inst{2} \and Thomas Besnier\inst{1} \and Stefan Sommer\inst{2}}
\institute{Univ. Lille, CNRS, Centrale Lille, UMR 9189 CRIStAL, F-59000 Lille, France \and Department of Computer Science (DIKU), University of Copenhagen, Denmark}

\begin{document}
\maketitle

\begin{abstract}
Modelling randomness in shape data, for example, the evolution of shapes of organisms in biology, requires stochastic models of shapes. This paper presents a new stochastic shape model based on a description of shapes as functions in a Sobolev space. Using an explicit orthonormal basis as a reference frame for the noise, the model is independent of the parameterisation of the mesh. We define the stochastic model, explore its properties, and illustrate examples of stochastic shape evolutions using the resulting numerical framework.
    
\end{abstract}

\section{Introduction}

%The recent rise of 3-dimensional data collection is accompanied by an increasing interest in the handling of such data. In particular, simulating processes depicting movements or any time-dependent evolution  is a challenging problem with crucial benefits for many applications such as human body modelling, evolutionary biology, or computer-generated imagery (CGI). 
In fields from medical imaging to biology, realistic models of shape must allow randomness in shape evolutions. For example, in evolutionary biology, random gene changes through evolution can be hypothesised to cause random shape variation. Therefore, there is a need for stochastic models of shape.
Moreover, recent years have seen the rise of diffusion models in deep learning \cite{Ho20,Dhariwal21} relying on a progressive addition of noise onto sample data which could motivate the search for new ways to define diffusion processes in particular data spaces such as shape spaces.
In this paper, we define a stochastic process for shape spaces that immerse a base shape in $\mathbb R^d$, $d=2,3$. The noise is related to the shape structure without reference to the ambient space in which the shape is embedded. Building on related models in the case of outer shape spaces, we construct a framework that is independent of the chosen shape representation and that gives numerically efficient ways of simulating stochastic shape evolutions.
%Our framework could be used as an alternative way to "add noise to 3D shapes" to build new diffusion models.

\subsection{Motivation and contribution}
%To avoid the complexity involved in processing voxels in 3D, we focus on shapes described explicitly in the \textit{shape space} through point clouds or meshes. Therefore, in all generality, a sampled shape $\overline{S}$, with $n$ degrees of freedom in the approximation of the shape, is a point in $\mathbb{R}^{3 \times n}$ for a point cloud, with additional connectivity information for a mesh. We seek to define random paths (stochastic processes) in this shape space starting from a source shape $S_0$ (and its discretisation $\overline{S}_0$). The major difficulty here resides in the shape space being a non-Euclidean, infinite-dimensional manifold. Recent work has been done in the case of landmark space through diffeomorphisms acting on the landmarks \cite{Sommer17,Arnaudon19,Arnaudon22}, and the next step is to find ways to define stochastics involving the whole shape. This paper explores an idea to do so through an intrinsic description of compact surfaces as functions. Thus, it defines a framework independent of the landmark configuration of the involved shapes.\\
We seek to define random paths (stochastic processes) in shape spaces starting from a source shape $S_0$. The major difficulty here resides in shape spaces generally being non-Euclidean, infinite-dimensional manifolds. Recent work has perturbed the momentum of Hamilton's equations \cite{Trouve12,Vialard13} or used stochastic perturbations defined in the ambient space \cite{Sommer17,Arnaudon19,Arnaudon22}. Here, we wish to define the stochastic perturbations directly in the shape space without  referring to the ambient space. This paper explores an idea to do so through an intrinsic description of compact surfaces as functions. Thus, it defines a framework independent of the discretisation of the shapes. However, a guiding principle for our work is that it should be computationally feasible to simulate from the constructed process. We demonstrate this with numerical simulations for which 
%\subsection{Contributions}
%This work describes and applies a new intrinsic framework to compute diffusion processes in a shape space structure. Concretely, this paper provides the following:
%\begin{itemize}
%    \item A mathematical model for diffusion processes in a meaningful shape space structure.
%    \item A publicly available implementation of numerical algorithms for intrinsic diffusion processes applied to 3D meshes.
%\end{itemize}
the code is accessible at %\url{*released-after-review*}. 
\url{https://github.com/tbesnier/bm-shapes}.
The method can efficiently integrate random shape trajectories. We highlight two distinct methods, one for shapes represented by functions and another for shapes represented explicitly by point clouds or meshes.

\subsection{Related Work}

This paper fits into a body of work on the analysis of shapes in shape spaces as follows. %To analyse shapes and, in particular, differences between shapes, we need a shape space. 
Shape spaces generally encompass definitions for shapes, paths between shapes, and lengths of such paths. 
There are multiple methods to define shape spaces \cite{Bauer14}; we mention two of them here. The first approach, used in this paper, identifies shapes as maps from an underlying manifold into $\bbbr^d,$ with $d=2,3$. Paths in the shape space are paths in the space of functions. In the second approach, shapes are considered as a subset of $\bbbr^d$. Variations of shapes arise from the action of the diffeomorphism group over $\bbbr^d$. %The diffeomorphisms are applied to the ambient space $\bbbr^d$, thereby acting on the shape. 
This approach leads to the \textit{Large Deformation Diffeomorphic Metric Mapping} (LDDMM) framework \cite{Younes2019,Beg05}. %Since the first approach applies directly to the shape; we refer to this as an \textit{intrinsic} approach. This contrasts with deforming an ambient space as in the LDDMM framework. We refer to this as an \textit{extrinsic} approach.
The metric structures appearing in the first class are often referred to as inner metrics. The second approach correspondingly leads to outer metrics.

\subsubsection{Inner approach}
Usually, the function space is taken to be the space of \textit{immersions} or \textit{embeddings}. Immersions are smooth maps for which the differential map is injective. An embedding is an injective immersion. Because embeddings are injective, it prevents the shape from self-intersecting. Different classes of metrics have been introduced in the literature to define distances between shapes represented as functions. 
One example is the square-root normal field (SRNF) \cite{Jermyn2012} in which shapes are elements of $L^2(\bbbs^2, \bbbr^3)$ and the $L^2$ distance is used to define a (pseudo-)distance on the space of immersions. Due to their simplicity and ease of calculation, SRNF has led to several numerical frameworks \cite{Laga17,Bauer19}. %The downside of using SRNF is that the defined metric can lead to undesirable behaviour, such as vanishing geodesics between different shapes \cite{Klassen2020,Bauer22}.
Other tools have been developed with stronger Sobolev metrics (also called elastic metrics) with additional theoretical properties \cite{Hartman22}. One example is from Su \textit{et al.} \cite{Su2020}: shapes are decomposed into a spherical harmonics basis, and a framework to find geodesics between the decomposed shapes is introduced. In this work, we will also
use spherical decompositions of shapes.

\subsubsection{Outer approach}
The LDDMM framework applies in a matching context from a source shape $S_0$ to a target shape $S_1$. An optimisation problem is solved to find the ``best'' diffeomorphism acting on $S_0$ and its ambient space to match $S_1$, resulting in geodesics in the shape space. A probabilistic framework can be built around this by considering the Hamiltonian formulation of the geodesic. That is, the geodesic equation is written in terms of a momentum and velocity equation, and these are perturbed. In \cite{Trouve12}, the momentum map is perturbed. This is executed in a finite setting where shapes are approximated by a finite number of points (called landmarks). Adding noise to the momentum equation can be interpreted as a random force acting on each landmark. In \cite{Vialard13}, the approach is extended to the case where the number of landmarks approaches infinity. This is similar to our approach, in that we too consider stochastic perturbations of maps in $L^2$-space. But, where \cite{Trouve12,Vialard13} perturb the momentum map of the geodesic equation, we perturb the shape directly by considering the shape as a function. 
More recent work \cite{Arnaudon19} has perturbed both the momentum and the velocity maps.

\section{Background}

\subsection{Shape space}
\subsubsection{Shapes as immersions and embeddings}
One way of modelling shapes is as functions from an underlying manifold $M$ into $\bbbr^d$. The shape space is usually taken as either the space of immersions or embeddings \cite{Bauer14}. The underlying manifold $M$ can be chosen based on the dimension and the topological features of the shape to be modelled \cite{Hartman22}. A common choice is $M = \bbbs^2$. Then, a shape $s$ is modelled as an immersion (or an embedding) $s:\bbbs^2 \rightarrow \mathbb{R}^3$. In this way, the shape $s$ deforms the sphere. In particular, $s$ belongs to $L^2(\bbbs^2, \mathbb{R}^3)$, the Hilbert space of square-integrable functions.

Taking shapes as immersions (resp. embeddings) gives parameterised shapes; two shapes that have identical images in $\bbbr^d$ with different parameterisations are treated as different shapes. We can also consider unparameterised shapes by taking the shape space to be the set of immersions (resp. embeddings) quotiented by the space of diffeomorphisms $\text{Diff}(M)$. Two shapes $s, s'$ are equivalent up to a reparameterisation if there exists a diffeomorphism $\phi \in \text{Diff}(M)$, such that $s = s' \circ \phi$. This forms an equivalence class $\text{Imm}(M, \bbbr^d)/\text{Diff}(M)$ over the space of immersions (resp. embeddings).
Therefore, the space of immersions (resp. embeddings) is called the \textit{preshape space}. Paths in the preshape space can be projected to the shape space via the projection map. Equivalently, the stochastic model defined in this paper can be mapped to the shape space by applying the projection to the process.

\subsubsection{Spherical harmonic decomposition of shapes}
\begin{figure}
    \centering
    \includegraphics[width=0.7\textwidth]{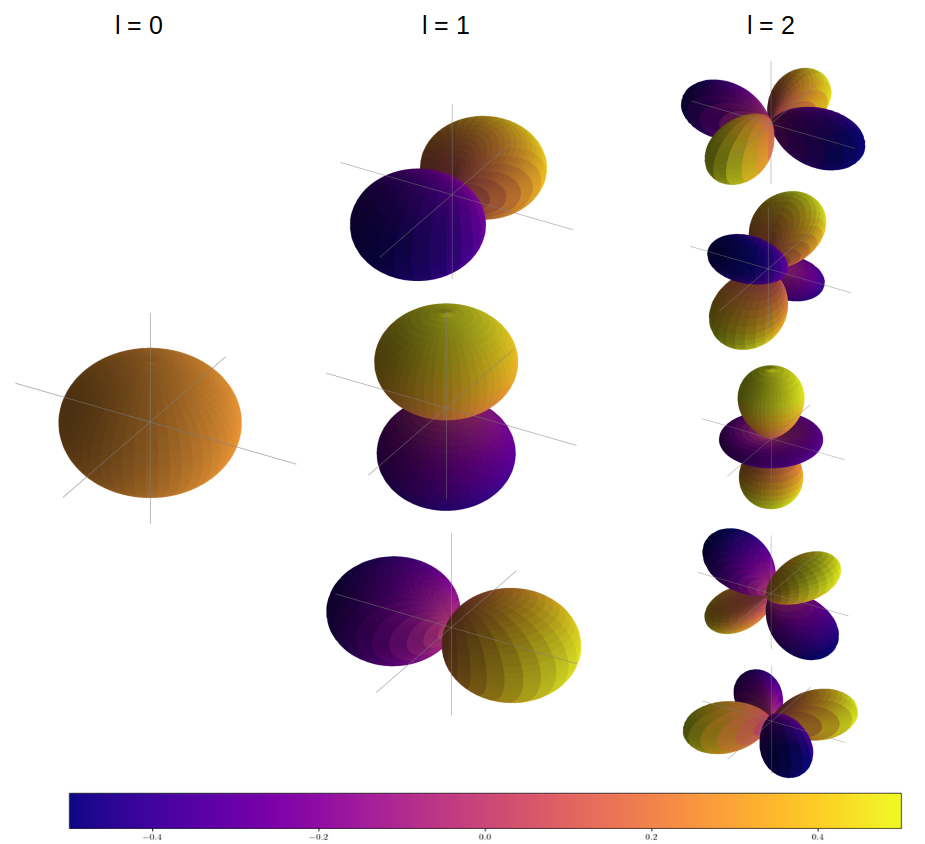}
    \caption{Visualisation of the three first orders of elementary functions in the spherical harmonic basis}
    \label{fig:sph_harm_viz}
\end{figure}
Considering a shape as an element of $L^2(\bbbs^2, \mathbb{R}^3)$ enables an orthonormal expansion of the shape. We choose spherical harmonics for the orthonormal basis as it is a natural choice for functions over the sphere %\cite{Reid86}.
\cite{Courant1989}; loosely speaking, it is the ``spherical counterpart'' of the Fourier decomposition.

Let $L^2(\mathbb{S}^2)$ be the space of square-integrable functions $f: \mathbb{S}^2 \to \mathbb{R}$, with $\bbbs^2$ parameterised by $\omega = (\theta, \phi) \in [0,2\pi) \times [0,\pi)$. Then $L^2(\bbbs^2)$ forms a Hilbert space when equipped with the inner product 
\begin{equation*}
    \langle f , g \rangle_{L^2(\bbbs^2)} := \int_{\mathbb{S}^2}f(\omega)g(\omega)\diff\omega =\int_0^\pi \int_0^{2\pi}f(\theta, \phi)g(\theta, \phi)\sin(\phi)\diff\theta \diff\phi,
\end{equation*}
for $f, g \in L^2(\bbbs^2).$

Let $\Delta_{\mathbb{S}^2}$ be the Laplace-Beltrami operator on the sphere. The \textit{spherical harmonics} $Y_l^m: \bbbs^2 \to \bbbr$ are defined as the eigenfunctions of $\Delta_{\mathbf{S}^2}$ with respective eigenvalues $-l(l+1)$:
$$-\Delta_{\mathbb{S}^2} Y_l^m = l(l+1) Y_l^m.$$
The spherical harmonics form a complete orthonormal basis of $L^2(\bbbs^2)$: any function $s \in L^2(\bbbs^2)$ can be written as
\begin{equation}
    s(\theta, \phi) = \sum_{l=0}^{\infty} \sum_{|m|\leq l}\langle s, Y_l^m \rangle_{L^2(\bbbs^2)} Y_l^m(\theta, \phi) = \sum_{l=0}^{\infty} \sum_{|m|\leq l}\hat{s}_{l,m} Y_l^m(\theta, \phi).
\end{equation}
The spherical harmonics have an explicit formula  for every $l \in \mathbb{N}, |m|\leq l$ as
\begin{equation}
    Y_l^m(\theta, \phi) = \sqrt{\frac{2l + 1}{4\pi} \frac{(l - m)!}{(l + m)!}} P_l^m(\cos(\theta))e^{im\phi}
\end{equation}
with $P_l^m$ being the \textit{associated Legendre polynomials}:
\begin{equation}
    P_l^m(x) = (-1)^m2^l(1-x^2)^{m/2} \sum_{k=m}^l\frac{k!}{(k - m)!}x^{k - m}\binom{l}{k}\binom{{(l+k - 1)}/{2}}{l}.
\end{equation}
%The preceding spherical harmonics basis $\{Y_l^m\}_{\substack{l \in \mathbb{N} \\ |m|\leq l}}$ is a complete orthonormal system for $L^2(\bbbs^2)$ and is analogous to the Fourier basis. 
The space $L^2(\bbbs^2, \bbbr^3)$ is defined as triplets of functions in $L^2(\bbbs^2)$. That is
$$L^2(\bbbs^2, \bbbr^3) := \{f=(f_1, f_2, f_3) \mid f_1, f_2, f_3 \in L^2(\bbbs^2)\}.$$ 
\begin{comment}
When equipped with the inner product 
$$\langle f, g, \rangle_{L^2(\bbbs^2, \bbbr^3)} := 
\langle f_1, g_1\rangle_{L^2(\bbbs^2, \bbbr)} + 
\langle f_2, g_2\rangle_{L^2(\bbbs^2, \bbbr)} + 
\langle f_3, g_3\rangle_{L^2(\bbbs^2, \bbbr)},$$
it is a Hilbert space. 
\end{comment}
The spherical harmonic decomposition can therefore be extended from functions in $L^2(\bbbs^2, \bbbr)$ to functions $f=(f_1, f_2, f_3)$ in $L^2(\bbbs^2, \bbbr^3)$ by taking the spherical harmonic decompositions of $f_1, f_2$ and $f_3$. 

Here we have only detailed the spherical harmonics basis. However, any orthonormal basis $\{e_i\}_{i=0}^\infty$ of $L^2(\bbbs^2)$ could be used. Another potential choice could be spherical wavelets \cite{Antoine08,Lessig08}. Then, each shape $S$ can be represented as the weighted sum of basis elements $S = \sum_{i=0}^\infty \alpha_i e_i$ for some coefficients $\alpha_i \in \bbbr$, and orthonormal basis $\{e_i\}_{i=0}^\infty$ of $L^2(\bbbs^2)$.\\
 
\subsubsection{Sobolev spaces}
Our aim is to use stochastic processes in the Hilbert space $L^2(\bbbs^2, \bbbr^3)$ to define stochastic evolutions of shapes. The advantage of working in the $L^2$ space is that there is an explicit basis. However, functions in $L^2(\bbbs^2, \bbbr^3)$ need not even be continuous, leading to highly irregular surfaces (see \cref{fig:sphere_process} for an example). To circumvent this, we will work in the Sobolev space $H^\nu(\bbbs^2, \bbbr^3)$, a subspace of $L^2(\bbbs^2, \bbbr^3)$.

The Sobolev space $H^{\nu}(\bbbs^2)$ of order $\nu \in \mathbb{N}_{\geq 0}$ is defined by

\begin{equation}
    H^\nu(\mathbb{S}^2) := \left\{ f \in L^2(\mathbb{S}^2) \; \middle| \;
    \sum_{l\in \bbbn}\sum_{|m|\leq l} 
    (l+1)^{2\nu}|\hat{f}_{l,m}|^2 < \infty
 \right\},
\end{equation}
where $\hat{f}_{l, m} = \langle f, Y_l^m \rangle_{L^2(\bbbs^2)}$ \cite{Gia2010}. This is a Hilbert space when endowed with the inner product 
$$\langle f, g \rangle_{H^{\nu}(\bbbs^2)}:= \langle f, g \rangle_{L^2(\bbbs^2)} + \langle (-\Delta_{\mathbb{S}^2})^{\nu/2}f, (-\Delta_{\mathbb{S}^2})^{\nu/2}g \rangle_{L^2(\bbbs^2)},$$
where $\Delta_{\bbbs^2}$ is the Laplace-Beltrami operator on the sphere, and the fractional power of the Laplace-Beltrami operator is defined in terms of spherical harmonics as

$$(-\Delta_{\mathbb{S}^2})^{\nu/2}f = 
\sum_{l\in \bbbn}\sum_{|m|\leq l}(l(l+1))^{\nu/2} \hat{f}_{l,m} Y_l^m.$$

 By the embedding theorem, functions in $H^\nu(\bbbs^2)$ are continuous for any $\nu\geq 2$. Essentially, functions in Sobolev spaces have fast-decaying spherical harmonic coefficients. In other words, their spectral information concentrates around low frequencies. Moreover, this gives us a method for mapping functions from $L^2(\bbbs^2)$ into $H^\nu(\bbbs^2)$:
If $$f= \sum_{l, m} \hat{f}_{l, m} Y_l^m\in L^2(\bbbs^2)$$ then $$g = \sum_{l, m} (l+1)^{-\nu}\hat{f}_{l, m} Y_l^m \in H^\nu(\bbbs^2).$$

\subsection{Stochastic processes in Hilbert spaces}
Throughout the rest of this paper, we assume our stochastic processes are defined over some probability space $(\Omega, \mathcal{F}, \bbbp).$

\subsubsection{Wiener processes in Hilbert spaces}
To define stochastic processes, % in $L^2(\bbbs^2, \bbbr^3)$, 
we apply the theory of stochastic processes in infinite dimensional Hilbert spaces \cite{DaPrato14}. For this, we discuss briefly how to define a $Q$-Wiener process $(W_t^Q)_{t \in [0, T]}$, with ending time $T>0$, in a Hilbert space. 

Let $\mathcal{H}$ be a Hilbert space and $Q$ a non-negative, trace-class operator on $\mathcal{H}$. Then, there exists some orthonormal basis $\{e_i\}_{i=0}^\infty$ of $\mathcal{H}$, and values $\lambda_i \in \bbbr$ such that
\begin{align}\label{eq: q wiener process}
    Qe_i = \lambda_i e_i, \; \text{for all } i \in \bbbn.
\end{align}
Define a $Q$-Wiener process as an $\mathcal{H}$-valued stochastic process
$$W^Q_t := \sum_{i=0}^\infty \sqrt{\lambda_i}B^i_te_i,$$
where $t\in [0, T]$ and $\{B^i_t\}_{i = 0}^\infty$ are independent, real-valued Brownian motions on the probability space $(\Omega, \mathcal{F}, \bbbp)$. The series \cref{eq: q wiener process} has the expected properties of a Wiener process: it converges in $L^2(\Omega, \mathcal{F}, \bbbp; C([0, T], \mathcal{H}))$ where $C([0, T], \mathcal{H})$ is equipped with the supremum norm, it has a continuous modification, and it has independent increments, with Gaussian laws \cite{DaPrato14}. %When $Q$ is the identity operator the process is called a \textit{cylindrical Wiener process} and is not defined in $H$.

\subsubsection{Real-valued Itô processes}
We want to compute stochastic processes over the coefficients of the spectral decompositions of shapes. To this end, we define an Itô process and state its convergence properties (see \cite{Schilling2014} for details). A (one dimensional) Itô process $(X_t)_{t \in [0, T]}$ is defined as the solution of a stochastic differential equation (SDE) of the form 
\begin{equation}\label{eq: ito process}
    \diff X_t = b(t, X_t) \diff t + \sigma(t, X_t)\diff B_t, \; X_0 \in L^2(\bbbp)
\end{equation}
where $B_t$ is Brownian motion on $\bbbr$ with respect to a filtration $(\mathcal{F}_t)_{t\in[0, T]}$ and $b: [0, T] \times \bbbr \to \bbbr$ and $\sigma: [0, T] \times \bbbr \to \bbbr$ satisfy the Lipschitz continuity condition:
$$|b(t, x) - b(t, y)|^2 + |\sigma(t, x) - \sigma(t, y)|^2 \leq L_T|x-y|^2,$$
where $x, y \in \bbbr, t \in [0, T]$ and $L_T< \infty$.
We call $b$ and $\sigma$ the \textit{drift} and \textit{diffusion} terms, respectively. The solution to \cref{eq: ito process} is unique for $\mathcal{F}_0$-measurable initial conditions $x_0 \in L^2(\bbbp)$ and satisfies
\begin{align}\label{eq: ito expectation}
\mathbb{E}\left[\sup_{t\in [0, T]} \|x_t\|^2 \right] \leq \kappa_T \cdot \mathbb{E}\left[ (1 + |x_0|)^2\right].
\end{align}
The choice of $b$ and $\sigma$ affects the behaviour of the Itô processes.

\section{Inner approach: Spectral diffusion}
Our aim is to introduce an inner shape space approach to stochastics in shape spaces. In order to develop stochastic evolutions of shapes, we represent shapes via a spherical harmonic decomposition. We can consider the stochastic evolution of shapes by adding a $Q$-Wiener process (or any diffusion process) directly to the decomposition and then constraining the process to $H^\nu(\mathbb{S}^2, \mathbb{R}^3), \; \nu\geq 2$. 

Given a shape $u_0 \in L^2(\bbbs^2, \bbbr^3)$ and an operator $Q$ on $L^2(\bbbs^2, \bbbr^3)$, we define a $Q$-Wiener process in the spectral domain of a shape, $u_t^Q$ as
\begin{equation}\label{eq: spectral_diffusion}
    u^Q_t = u_0 + W_t^Q = \underbrace{\sum_{l=0}^\infty \sum_{|m|<l} \langle u_0, Y_l^m \rangle Y_l^m}_\text{source shape} + \underbrace{\sum_{l=0}^\infty \sum_{|m|<l}B_t^{l, m} Q^{1/2} (Y_l^m)}_\text{$Q$-Wiener process}
\end{equation}
where $\{B^{l, m}_t\}_{l,m}$ are independent real-valued Brownian motions.

In \cref{eq: spectral_diffusion}, $Q$ controls the diffusion to guarantee the convergence of the process in $H$. If we choose $Q$ to be a non-negative trace-class operator, the process converges in $L^2(\bbbs^2, \bbbr^3)$. For any basis $\{e_i\}_{i\in \bbbn}$ of $L^2$ and any sequence of positive real numbers $\{\lambda_i\}_{i\in \bbbn}$ satisfying $\sum_{i\in \bbbn} \lambda_i < \infty$, the operator 
$$Q(\cdot) = \sum_{i\in \bbbn} \lambda_i \langle \cdot, e_i\rangle e_i$$
is non-negative and of trace class. Therefore, we use a spherical harmonics basis weighted with a positive decaying sequence $\{\lambda_i\}_{i\in \bbbn},$ where $\sum_i \lambda_i < \infty,$ defining the $Q$-Wiener process
$$W_t^Q = \sum_{l\in \bbbn}\sum_{|m|\leq l}\sqrt{\lambda_l} B^{l, m}_t Y_l^m.$$
The rate of decay of the coefficients $\{\lambda_i\}_i$ controls the regularity of the space in which the process converges. If 
$$\sum_{l\in \bbbn}\sum_{|m|\leq l} \lambda_l (l+1)^{2\nu} < \infty$$
the sum converges in $L^2(\Omega; C([0, 1], H^\nu(\bbbs^2, \bbbr^3)),$ where $C([0, 1], H^\nu(\bbbs^2, \bbbr^3))$ is equipped with the supremum norm. 

The process in \cref{eq: spectral_diffusion} can be generalised by exchanging the independent Brownian motions $\{B^{l, m}_t\}_{l, m}$ with other, more general independent stochastic processes $\{x_t^{l, m}\}_{l, m}$, for example, Itô processes. In this way, we have stochastic processes on the coefficients of the shape. Letting $x^{l, m}_0 = 0$ means that \cref{eq: ito expectation} is bounded. When this is satisfied, the process
\begin{align}\label{eq: Ito coefficient diffusion}
    X_t^Q := \sum_{l\in \bbbn}\sum_{|m|\leq l}\sqrt{\lambda_l} x^{l, m}_t Y_l^m
\end{align}
also converges in $L^2(\Omega, \mathcal{F}, \bbbp; C([0, 1], H^\nu(\bbbs^2, \bbbr^3))$ \cite{DaPrato14}.

\begin{comment}
\begin{figure}
    \centering
    \includegraphics[width=0.95\textwidth]{img/projection_H2.PNG}
    \caption{The operator $Q$ acts as a projection operator onto $H^2$ ensuring desirable spatial regularities and containing the noise to ensure convergence.}
    \label{fig:projection_H2}
\end{figure}
\end{comment}

%In this section, we have presented our approach for defining stochastics in a shape space. To do so, we represent a shape as a function in $L^2(\bbbs^2, \bbbr^3)$ and decompose it in terms of the spherical harmonic basis. Then, we use the spherical harmonic decomposition to define a stochastic process in $L^2(\bbbs^2, \bbbr^3)$ with the initial value given by the shape. We constrain the stochastic process to $H^\nu(\bbbs^2, \bbbr^3)$ by weighting the coefficients of the decompositions.

\section{Numerical experiments for spectral diffusions}

We here aim to illustrate the constructed stochastic process. We address two situations. The first assumes shapes are provided via a function in its spherical harmonic decomposition, and the second assumes shapes are represented by meshes (with vertex coordinates and face connectivity information).
In the first situation, all that remains is to simulate the real-valued stochastic process $x^{l, m}_t$ from \cref{eq: Ito coefficient diffusion}. For the numerical integration, we use an Euler--Maruyama scheme \cite{Kloeden2011} such that the stochastic process $u_t$ defined by
\begin{equation}\label{eq: sde}
    \diff u_t = b(t, u_t)\diff t + \sigma(t, u_t)\diff W_t
\end{equation}
is approximated by
\begin{equation}
    u_{t_{k+1}} \approx u_{t_k} + b(t_k, u_{t_k})\Delta t_k + \sigma(t_k, u_{t_k})\Delta W_{t_k}
\end{equation}
with $\Delta t_k = t_{k+1} - t_k$ and $\Delta W_{t_k} = W_{t_{k+1}} - W_{t_k} \sim \mathcal{N}(0, \sqrt{t_{k+1} - t_k})$.

\subsection{Effects of truncation and covariance operators}

\begin{figure}[ht]
    \centering
    \includegraphics[width=0.95\textwidth]{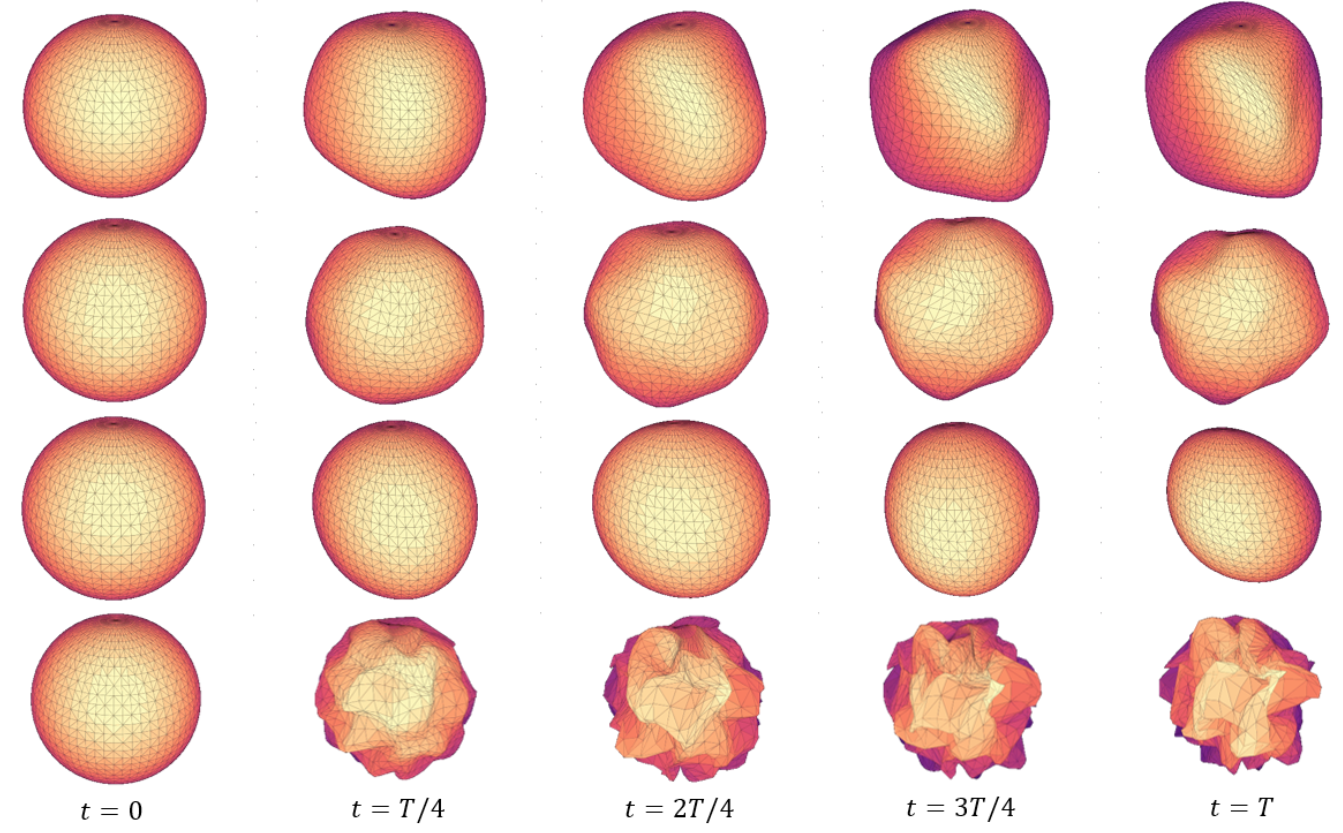}
    \caption{Each row presents some frames of a $Q$-Wiener process starting from the unit (discretised) sphere. 1st row: $\lambda_l=1$ up to $N=25$ (no decay), 2nd row: $\lambda_l = \frac{1}{l + 1}$, 3rd row: $\lambda_l = \frac{1}{(l+1))^2}$ and 4th row: $\lambda_l = 1$ up to $N=225$ (no decay). We highlight the loss of regularity in the last row when high spectral orders are not weighted down.}
    \label{fig:sphere_process}
\end{figure}

We start by illustrating the effects of different covariance operators within our framework. To this end, we use the sphere $\bbbs^2$, which we view as the image of
\begin{align*}
    s: \begin{cases}[0, \pi) \times [0, 2\pi) & \to \bbbr^3\\
    (\theta, \phi) & \mapsto (\sin{\theta}\cos{\phi}, \sin{\theta}\sin{\phi}, \cos{\theta}), 
    \end{cases}
\end{align*}
and then decompose $s$ with respect to the spherical harmonics basis
$$s= \left( 
\sum_{\substack{l\in \bbbn \\ |m| \leq l}} s^1_{l, m} Y_l^m, \; \sum_{\substack{l\in \bbbn \\ |m| \leq l}} s^2_{l, m} Y_l^m,\;
\sum_{\substack{l\in \bbbn \\ |m| \leq l}} s^3_{l, m}Y_l^m
\right).$$
The results in \cref{fig:sphere_process} show some processes $u(t) = (u^1(t), u^2(t), u^3(t))$ defined by
$$u^i(t) = s + \sum_{l=0}^N\sum_{|m|\leq l} \lambda_l B_t^{l, m}Y_l^m\; \in L^2(\bbbs^2, \bbbr) , \quad i\in \{1,2,3\}$$
where $N\in \bbbn$, $\{B_t^{l, m}\}_{l,m}$ are independent real-valued Brownian motions and $\{\lambda_l\}_{l \leq N}$ are chosen to define various operators. We also note the role the truncation value $N\in \bbbn$ plays. The spectral information of the series is concentrated around lower frequencies. Taking lower values for $N$ results in notably smoother shapes but also gives less detail to the surface of the shapes. We encourage the reader to also look at the .gif files of the simulations in the GitHub repository %\url{*released-after-review*}.
\url{https://github.com/tbesnier/bm-shapes}.

\subsection{Simulations on radial projections of meshes}

\begin{figure}[ht]
    \centering
    \includegraphics[width=1.0\textwidth]{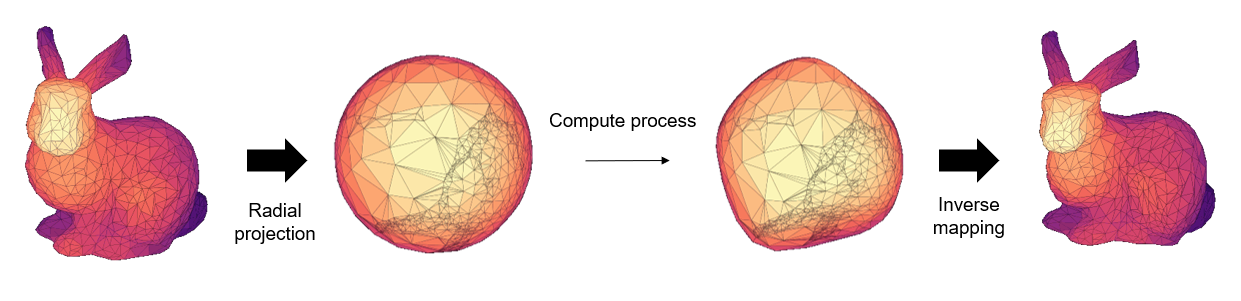}
    \caption{Numerical framework of our method on mesh data. First, we project on the sphere (with a radial projection, for instance), then, we compute the diffusion on the sphere and transfer the diffusion back to the mesh.}\label{fig:numerical_framework_general_shape}
\end{figure}

\begin{figure}[ht]
    \centering
    \includegraphics[width=1.0\textwidth]{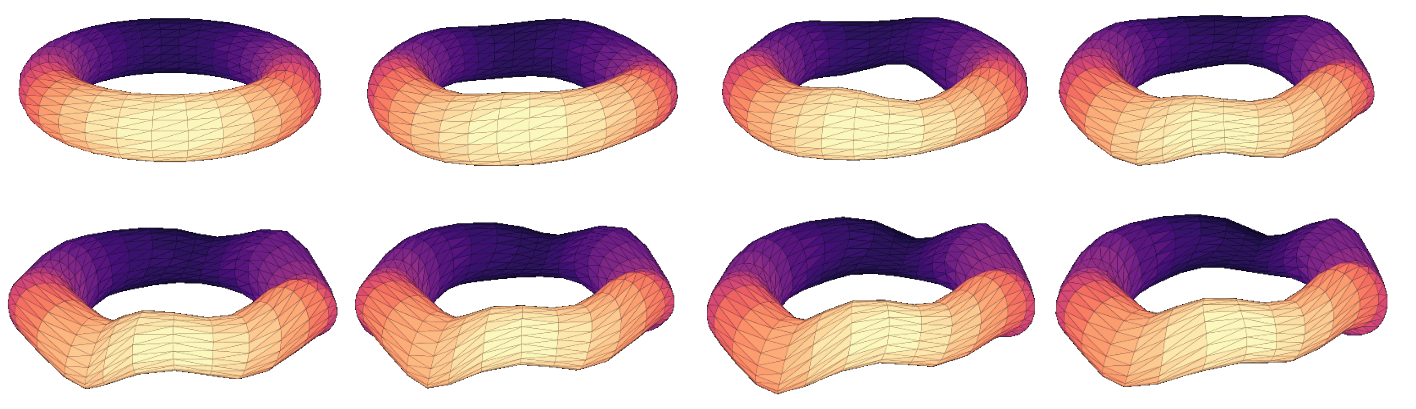}
    \caption{Simulation of a $Q$-Wiener process with $Q = \mathrm{Id}_{49}$. We show in this example how the spatial correlation prevents the sides of the torus from intersecting each other.}
    \label{fig:torus_process}
\end{figure}

3D shape data can be explicitly represented as meshes rather than as maps. Some work \cite{Kurtek2013,Praun2003} has been done on converting from mesh representations to spherical harmonic representations. In this section, we use an easier method using radial projection (see \cref{fig:numerical_framework_general_shape} for a visualisation) before computing a diffusion process from the sphere as described before. Finally, we take the inverse radial projection mapping of the deformed sphere.
We used this setting because of its simplicity regarding computations, but other types of spherical parameterisation of meshes \cite{Wang14,Asirvatham05} can be used. When using radial projections, the topology of the shape is no longer a problem since shapes of any genus can be projected onto the sphere. However, using the radial projection in this way means points with similar angular coordinates are highly correlated. For example, on the torus, it means that the outer and inner rings behave similarly (see \cref{fig:torus_process}). %In general, however, this behaviour may not be desired

\begin{figure}[ht]
    \begin{center}
    \includegraphics[width=1.0\textwidth]{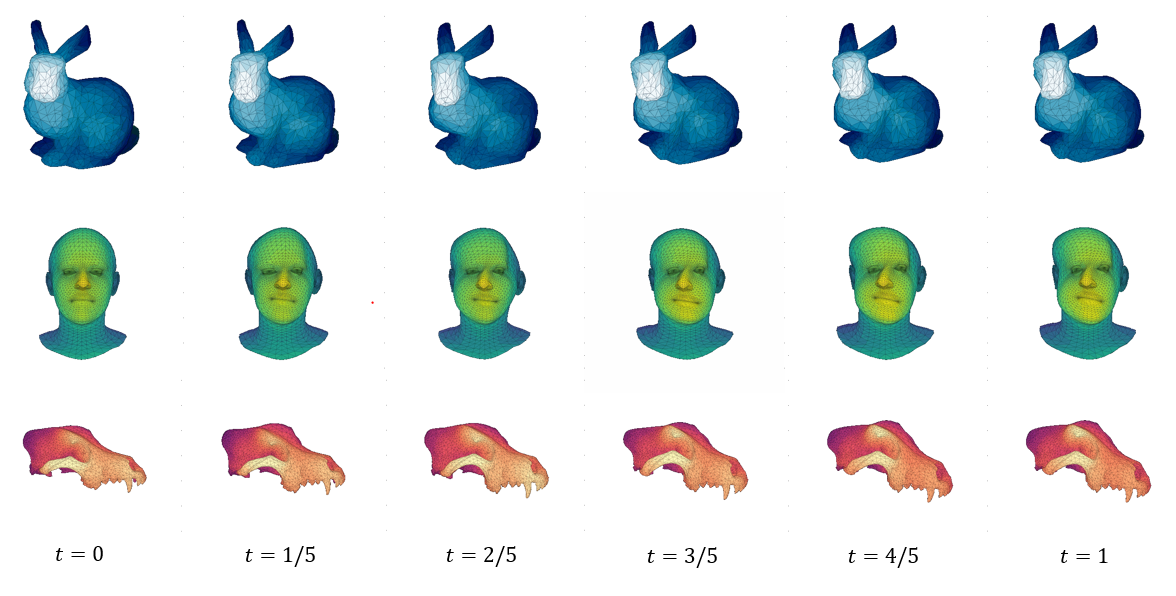}
    \end{center}
    \caption{Q-Wiener processes applied on different meshes. Here, $Q = (1 - \Delta_{\bbbs^2})^{-1}$ and we take the first 25 coefficients.}
    \label{fig:Q_Wiener_general_shapes}
\end{figure}

In \cref{fig:Q_Wiener_general_shapes} we illustrate our framework with radial projections for assorted meshes.

\subsection{Other types of processes}

\begin{figure}[ht]
    \begin{center}
    \includegraphics[width=0.95\textwidth]{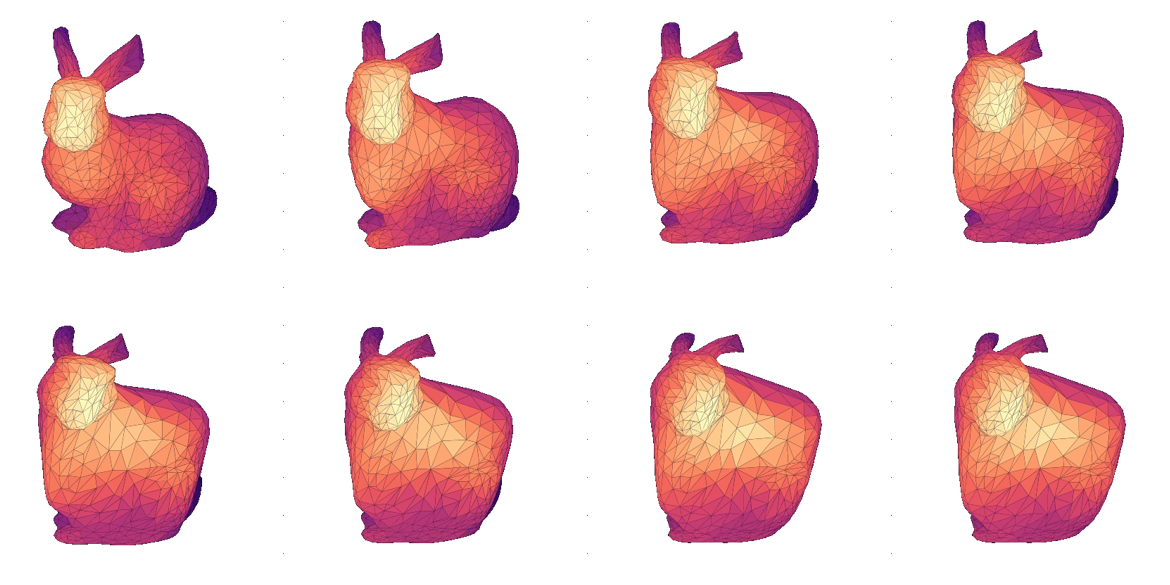}
    \end{center}
    \caption{Simulation of an Ornstein-Uhlenbeck diffusion, solution to an SDE with a drift term $b(t,x) = x$ and diffusion term $\sigma(t,x) = 0.1$. We highlight that, as expected, the process converges towards the sphere.}
    \label{fig:OU_process}
\end{figure}

%As we can compute Wiener processes in Hilbert spaces, we can also define and simulate any diffusion processes, unique solutions of an SDE of the form described by \cref{eq: sde}. 
We can define and simulate any diffusion processes, as in \cref{eq: Ito coefficient diffusion}, instead of using Brownian motion.
For instance, the Ornstein-Uhlenbeck process is defined by an SDE with the drift and diffusion terms given by $b(t,x):= x$ and $\sigma(t,x) := C$ respectively, where $C$ is a real constant. We observe in \cref{fig:OU_process} how the mesh inflates into a sphere which is expected for this type of process.
\begin{comment}
We next consider more general shapes. For these experiments, we use the data provided by \cite{Su2020}, represented directly by a spherical harmonic decomposition with resolution ??. In figure X, we simulate bridges between shapes via Brownian bridges on the coefficients. That is for two shapes $s_0 = (s_0^1, s_0^2, s_0^3), s_1 = (s_1^1, s_1^2, s_1^3)$, define the stochastic processes $x^i_{l, m}$ from \cref{eq: spectral_diffusion} via the SDE
$$\mathrm{d}x^i_{l, m}(t) =
\frac{(\beta^i_{l, m}- \alpha^i_{l, m}) - {x}^i_{l, m}(t)}{T-t} \mathrm{dt} + \mathrm{d}W_t, \qquad x^i_{l, m}(0) = 0.$$
Then $x_{l, m}(0)=0$ and $x_{l, m}(T) = \beta^i_{l, m}- \alpha^i_{l, m}$ almost surely.
\end{comment}
~\

In case a different time covariance structure is needed, we can also simulate \textit{fractional $Q$-Wiener processes}, which are generalisations of the Wiener process with a covariance function $C_h(s,t) = \frac{1}{2}(|t|^h + |s|^h - |t - s|^{2h})$ with Hurst index $h \in (0,1)$. The covariance function of Brownian motion corresponds to $h = 0.5$.
\begin{figure}[ht]
    \centering
    \includegraphics[width=0.95\textwidth]{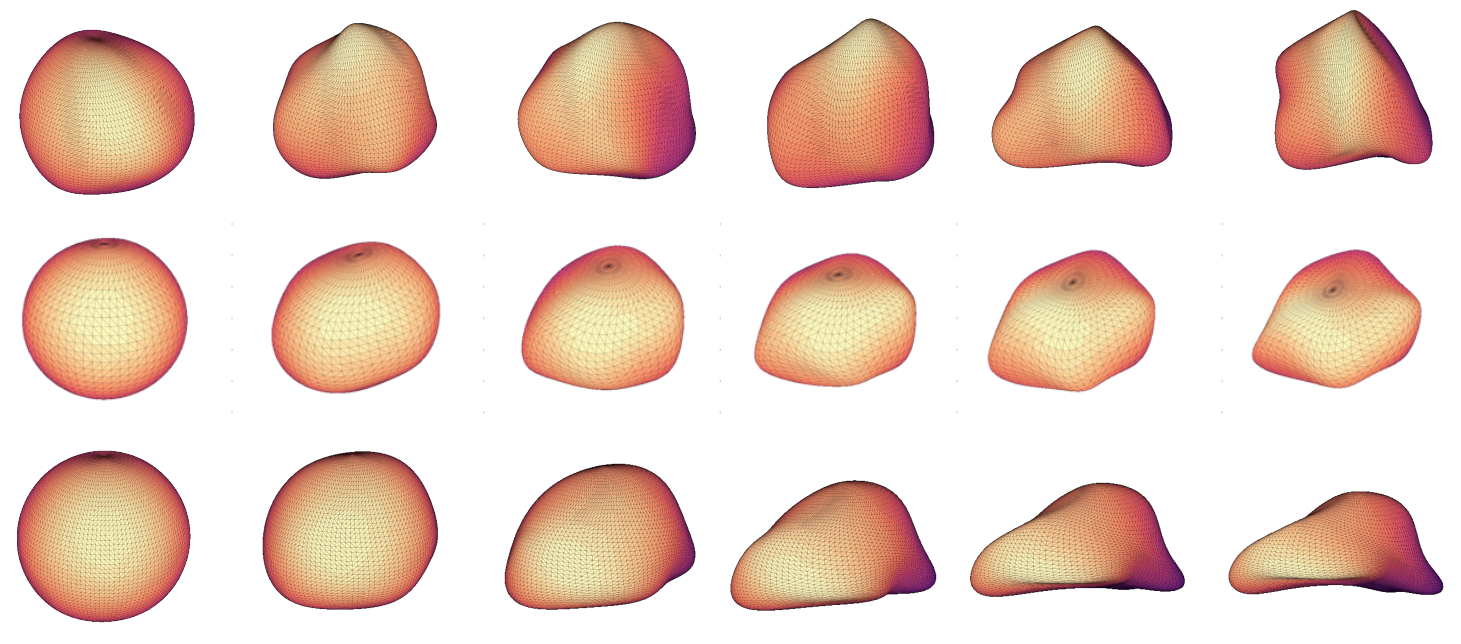}
    \caption{Simulation of a fractional Wiener process from the sphere with different Hurst indices: 1st row $h = 0.3$, 2nd row $h = 0.7$, 3rd row $h = 0.95$. The processes are simulated with the method described in \cite{Coeurjolly18}. As $H$ increases, the mesh is less subject to strong variations over short time intervals.}
    \label{fig:fractional_0-7}
\end{figure}
A whole theory is derived from this process, and we invite the interested reader to read \cite{Jumarie00} for more details. If $h<0.5$ or $h>0.5$, the process has negatively or positively correlated increments. We show resulting paths in \cref{fig:fractional_0-7} with different Hurst indices.

\section{Limitations and future work}

\subsection{Precision of the numerical reconstruction}
\begin{figure}[ht]
    \centering
    \includegraphics[width=1.0\textwidth]{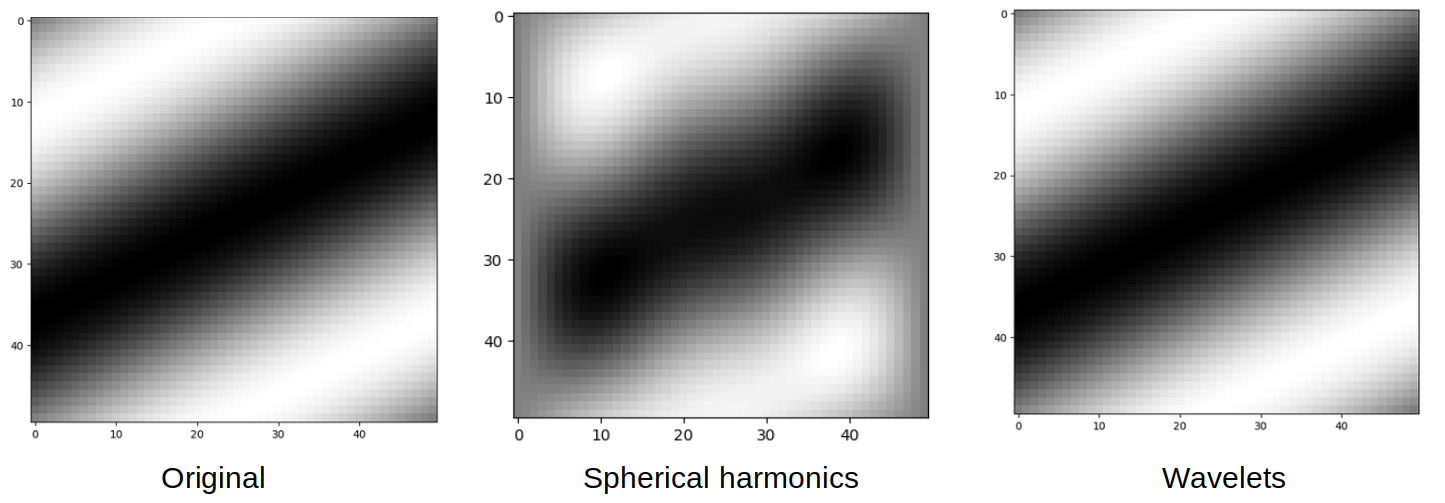}
    \caption{Reconstruction of a (projected) spherical signal. The original signal (on the left) is hardly reconstructed with spherical harmonics up to order 5. On the right, also with a resolution of 5, the wavelet (symlet 8 in \cite{Daubechles06}) reconstruction shows near-perfect results.}
    \label{fig:recons_error_sph}
\end{figure}

Decomposing a mesh with spherical harmonics is one of many choices for a basis of $L^2(\bbbs^2, \mathbb{R}^3)$. Wavelets could be better suited if the signal has sharp, irregular details or discontinuities (as demonstrated in~\cref{fig:recons_error_sph}) \cite{strang93}. But, the relation between Sobolev spaces and spherical wavelets is not as straightforward as for spherical harmonics.

\subsection{Limitations regarding the control of the process}

If we only have mesh data, the diffusion process is computed on the sphere and then mapped back to the mesh. It has the advantage of keeping spatial coherence as described before, but the spectral diffusion process has an isotropic spatial variance because of the symmetry properties of spherical harmonics. It becomes challenging to localise larger variance on some regions of the mesh, which could be desirable as one can assume changes in the variability of some areas with morphological data.

Finally, we point out that the sequence of weighting coefficients $\{\lambda_l\}_{l\in \mathbb{N}}$ is fixed and comes from the Laplace-Beltrami operator on the sphere $\mathbb{S}^2$. Rigorously, to apply the Bessel potential, we should recompute the eigenvalues of the Laplacian at each time step to stay in the shape space. We chose to avoid this extra step as it is a computationally costly operation, but it can be added in further developments.

\section{Conclusion}
In this paper, we proposed a function-based approach to compute stochastic processes (Itô diffusions) between discretised surfaces (effectively meshes) in $\mathbb{R}^3$. We can compute stochastics independently of the underlying parameterisation by modelling the mesh through its spherical harmonic decomposition. In addition, restraining the process to a Sobolev space ensures spatial regularity at all times. Our framework enables us to use various stochastic processes and visualise their behaviour in the mesh space with publicly available code.
Testing the behaviour of our framework on more complex 3D structures with different stochastic processes and applying it to geometry processing tasks is future work.

\subsubsection{Acknowledgements}
The work presented in this article was done at the Center for Computational Evolutionary Morphometry and is partly supported by Novo Nordisk Foundation grant NNF18OC0052000 as well as VILLUM FONDEN research grant 40582 and UCPH Data+ Strategy 2023 funds for interdisciplinary research.

\bibliographystyle{plain}
\bibliography{bib}

\end{document}